\documentclass{article}
\usepackage[utf8]{inputenc}
\usepackage{graphicx}
\usepackage{framed,multirow}
\usepackage[letterpaper, portrait, margin=1in]{geometry}

\providecommand{\keywords}[1]
{
  \small	
  \textbf{\textit{Keywords---}} #1
}

\title{Suspicious Behavior Detection on Shoplifting Cases for Crime Prevention by Using 3D Convolutional Neural Networks.}
\author{
  Martínez-Mascorro, Guillermo A.\\
  \texttt{a00824126@itesm.mx}
  \and
  Abreu-Pederzini, José R.\\
  \texttt{a00793921@itesm.mx}
  \and
  Ortiz-Bayliss, José C.\\
  \texttt{jcobayliss@tec.mx}
  \and
  Terashima-Marín, Hugo\\
  \texttt{terashima@tec.mx}
}
\date{}

\begin{document}

\maketitle

\begin{abstract}
    Crime generates significant losses, both human and economics. Every year, billion of dollars are lost due to attacks, crimes, and scams. Surveillance video camera networks are generating vast amounts of data, and the surveillance staff can not process all the information in real-time. The human sight has its limitations, where the visual focus is among the most critical ones when dealing with surveillance. A crime can occur in a different screen segment or on a distinct monitor, and the staff may not notice it. Our proposal focuses on shoplifting crimes by analyzing special situations that an average person will consider as typical conditions, but may lead to a crime. While other approaches identify the crime itself, we instead model suspicious behavior ---the one that may occur before a person commits a crime---  by detecting precise segments of a video with a high probability to contain a shoplifting crime. By doing so, we provide the staff with more opportunities to act and prevent crime. We implemented a 3DCNN model as a video feature extractor and tested its performance on a dataset composed of daily-action and shoplifting samples. The results are encouraging since it correctly identifies 75\% of the cases where a crime is about to happen.
\end{abstract}


\section{Introduction}
\label{sec:introduction}
    According to the 2018 National Retail Security Survey~(NRSS)~\cite{NRSS2018} inventory shrink, a loss of inventory related to theft, shoplifting, error or fraud, had an impact of \$46.8 billion in 2017 on U.S. retail economy. A high number of scams occur every day, from distractions and bar code-switching to booster bags and fake weight strategies, and there is no human power to watch every one of these cases.
    
    The surveillance context is overwhelmed. Vigilance camera networks are generating vast amounts of video screens, and the surveillance staff cannot process all the available information. The more recording devices become available, the more complex the task of monitoring such devices becomes.
    
    Real-time analysis of each camera has become an exhaustive task due to human limitations. The primary human limitation is the Visual Focus of Attention (VFOA) \cite{VisualFocusOfAtenttion}. Human gaze can only concentrate on one specific point at once. Although there are large screens and high-resolution cameras, a person can only regard a small segment of the image at a time. Optical focus is a significant human-related disadvantage in the surveillance context. A crime can occur in a different screen segment or on a different monitor, and the staff may not notice it. Other significant difficulties can be the attention paid, boredom, distractions, lack of experience, among others~\cite{ComplexHumanActivities2011, OperatorPerformance2012}.
    
    \begin{figure}[htb]
        \centering
        \includegraphics[width = 0.98\textwidth]{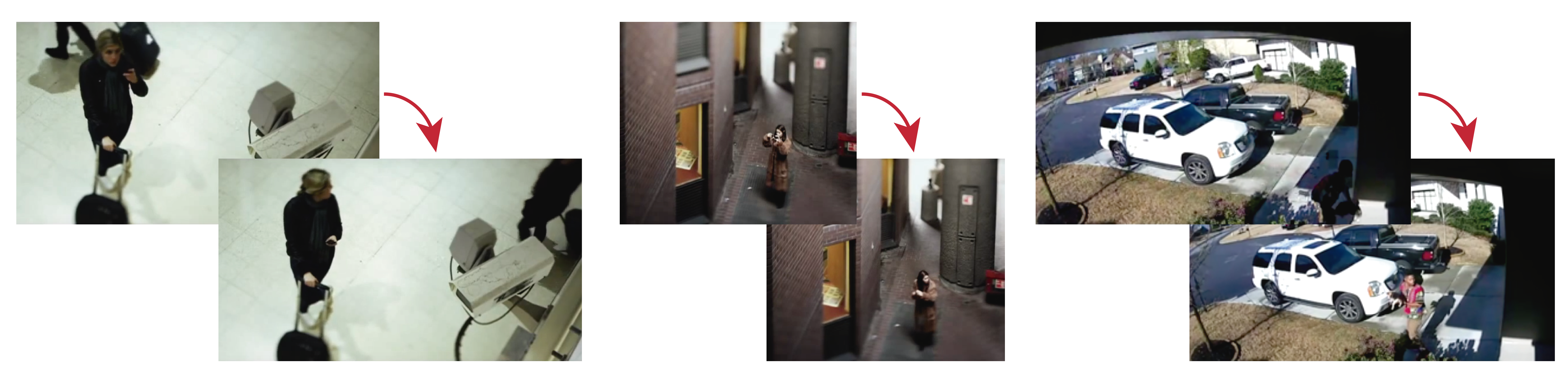}
        \caption{Suspicious behavior is not the crime itself, particular situations will make us distrust of a person.}
        \label{fig:SuspiciousBehavior}
    \end{figure}
    
    Defining what can be considered suspicious behavior is usually tricky, even for psychologists. In this work, the mentioned behavior is related to the commission of a crime, but it does not imply its realization (Figure \ref{fig:SuspiciousBehavior}). We define suspicious behavior as a series of actions that happen before a crime occurs. In this context, our proposal focuses on shoplifting crime scenarios, particularly on before-offense situations, that an average person may consider as typical conditions.
    While existing models identify the crime itself, we model suspicious behavior as a way to anticipate the crime. In other words, we identify behaviors that usually take place before a shoplifting crime. This kind of crime usually occurs in supermarkets, malls, retail stores, and other similar businesses. Many of the models for addressing this problem need the suspect to commit a crime to detect it. Examples of such models include face detection of previous offenders~\cite{FaceFirstSite, DeepCamSite} and object analysis in fitting rooms~\cite{SymmetryAnalysis}. In this work, we propose an approach that aims at supporting the monitoring staff to focus their attention on specific screens where crime is more likely to happen. By detecting situations in a video that may indicate that a crime is about to occur, we give the surveillance staff more opportunities to act, prevent, or even respond to such a crime. In the end, it is the security personnel who will decide how to proceed for each situation.
    
    We implement a 3D Convolutional Neural Network~(3DCNN) to process criminal videos and extract behavioral features to detect suspicious behavior. We perform the model training by selecting specific videos from the UCF-Crimes dataset~\cite{UCFCrimes2018}. Among the main contributions of this work, we propose a method to extract segments from videos that feed a model based on a 3DCNN and learns to classify suspicious behavior. The model achieves an accuracy of 75\% on suspicious behavior detection before committing a crime on a dataset composed of daily-action samples and shoplifting samples. These results suggest that our approach is useful for crime prevention in shoplifting cases.
        
    The remainder of this document is organized as follows. In Section~\ref{sec:Background}, we review various approaches that range from psychology to deep learning, to tackle behavior detection. Section~\ref{sec:Methodology} presents the methodology followed throughout the experimental process. The results and discussions of the tests are presented in Section~\ref{sec:Experiments}. Finally, Section~\ref{sec:Conclusion} presents the conclusions and future works derived from this investigation.
    
\section{Background and Related work}
\label{sec:Background}

    Every surveillance environment must satisfy with a particular set of requirements. Those requirements have promoted the creation of specialized tools, both on equipment and on software, to support the surveillance task. The most common approaches include motion detection~\cite{LOBS2012, GPUimplementation2013}, face recognition~\cite{FaceRecognition2015, FaceRecognition2018, DeepCamSite, FaceFirstSite}, tracking~\cite{ColourTracking2019, FnRTracking2011, HumanTracking2016}, loitering detection~\cite{Loitering2014}, abandoned luggage detection~\cite{localizedChang2010}, crowd behavior~\cite{crowdEscapeBehavior2014, DominantSets2014, abnormalWang2018}, and abnormal behavior~\cite{Ouivirach2012AutomaticSB, FastAccurateDetection2017}. 
    
    Prevention and reaction are two primary aims in the surveillance context. Prevention requires to forestall and deter a crime execution. The monitoring staff must remain alert, watching as much as they can, and alerting the ground personnel. Reaction, on the other hand, involves protocols and measures to respond to a specific event. The security teams take action only after the crime or event has taken place.
    
    Most security-support approaches focus on crime occurrence. \cite{SnatchingBehavior2019} present a snatching-detection algorithm, which performs background subtraction and pedestrian tracking, in order to make a decision. That approach divides the frame into eight areas and searches for a speed-shift in one of the tracked persons. The algorithm proposed by \cite{SnatchingBehavior2019} can only alert when a person already loses its belongings. \cite{UCFCrimes2018} present a real-world anomaly detection approach, training thirteen anomalies, such as burglary, fighting, shooting, and vandalism. They label the samples into two categories: normal and anomalous, and use a 3DCNN for feature extraction. Their model includes a ranking loss function and trains a fully-connected neural network for decision making. \cite{LoiteringZin2019} propose a system to detect loitering people. The system combines several analyzes for decision fusion and final detection, including distance, acceleration, direction-based, and grid-based analysis.
    
    Convolutional Neural Networks (CNN) have shown a remarkable performance in computer vision and different areas in the last recent years. Particularly, 3DCNN ---an extension of CNN---, focus on extracting spatial and temporal features from videos. Traditional applications that have been implemented using 3DCNN include object recognition~\cite{HeObject3DCNN2017}, human action recognition~\cite{3dCNNHumanAction2013}, gesture recognition~\cite{ZhangGestures3DCNN2017}, and as a specific implementation, Cai et al.~\cite{CaiAbnormal3DCNN2016} used 3DCNN for abnormal behavior detection in examination surveillance within the classroom. 
    
    Although all the works mentioned before are based on 3DCNN, each one has a particular architecture, and many parameters ---such as depth, number of layers, number of filters on each layer, kernel size, padding, stride--- must be adjusted. For example, concerning the number of layers, many approaches rely on simple structures that consist of two or three layers~\cite{3dCNNHumanAction2013, CaiAbnormal3DCNN2016, HeObject3DCNN2017}, while others require several layers for exhaustive learning~\cite{ZhangGestures3DCNN2017, VarolActionsCLSTM2018, KrizhevskyAlexNet2012, SzegedyGoogleNet2015, SimonyanVGGNet2014, HeResNet2015}.
    
    Concerning shoplifting, the current literature is somewhat limited. Surveillance material is, in most cases, a company's private property. The latter restricts the amount of data we can get to train and test new surveillance models. For this reason, several approaches focus on training to detect normal behavior. Anything that lies outside the cluster is considered abnormal. In general, surveillance videos contain only a small fraction of crime occurrences. Then, most of the videos in the data are likely to contain normal behavior.
    
    Many approaches have experienced problems regarding the limited availability of samples and their unbalanced category distribution. For this reason, some works have focused on developing models that learn with a minimal amount of data. For example,~\cite{3dCNNHumanAction2013} create a school dataset and test with eight to ten videos,~\cite{SnatchingBehavior2019} rely on a dataset of nineteen videos (four used for training and fifteen for testing), and~\cite{LoiteringZin2019} work with six videos (one for training and five for testing). 
    
    This work aims at developing a support approach for shoplifting crime prevention. Our model detects a person that, according to its behavior, is likely to commit a shoplifting crime. We achieve the latter by analyzing the comportment of the people that appear in the videos before the crime occurs. To the best of our knowledge, this is the first work that analyzes behavior as a means to anticipate a potential shoplifting crime.

\section{Methodology}
\label{sec:Methodology}

As part of this work, we propose a new methodology to extract segments from videos where people exhibit behaviors that are relevant to the task of preventing shoplifting crime. These behaviors include both normal and suspicious, being the task of the network to classify them. In this section, we will describe the dataset used for experiments and the 3DCNN architecture used for feature extraction and classification.
    
    \subsection{Description of the Dataset}
    \label{sec:Dataset}        
       
    We use the UCF-Crime dataset, proposed by~\cite{UCFCrimes2018}, to analyze suspicious behavior before a shoplifting crime. The dataset consists of 1900 real-world surveillance videos and provides around 129 hours of videos (with a resolution of 320x240 pixels and not normalized in length). The dataset includes scenarios from several locations and persons that are grouped into thirteen classes such as `abuse', `burglary', and `explosion', among others.  From those classes, we extracted the samples used in this work from the `shoplifting' and `normal' classes.
    
    To feed our model, we require videos that show one or more persons and that their activities are visible before the crime is committed. Because of these restrictions, not all the videos in the dataset are useful. Suspicious behavior samples were extracted only from videos that exhibit a shoplifting crime ---and these samples omit the crime itself. Normal behavior samples were taken from the `normal' class. Thus, it is important to stress that the model we propose is a suspicious behavior classifier and not a crime classifier.
    
    For processing the videos and extracting the suspicious samples (video segments that exhibit a suspicious behaviour), we propose a new method, the Pre-Crime Behavior (PCB) analysis, which we explain in the next section. Once we obtain the suspicious samples, we applied some transformations to produce several smaller datasets. First, to reduce the computational resources required for training, all the frames in the videos were transformed into grayscale and resized to four testing resolutions: 160$\times$120, 80$\times$60, 40$\times$30, and 32$\times$24 pixels. For organization purposes, all the samples extracted from the videos are indexed. The suspicious samples are indexed as SB$_{i}$ (where $i$ ranges from 1 to 60) while the samples of normal behavior are indexed as NB$_{i}$ (where $i$ ranges from 1 to 60). We divided the original sample size by 2, 4, 8, and 10 to explore the performance of each configuration. Table~\ref{tab:SamplesSelection} describes how these datasets are conformed. For example, \textit{SBT\_balanced\_120} is a set that contains 120 samples with the same number of suspicious and normal samples, 60 of each class (samples SB$_{1}$ to SB$_{60}$ and NB$_{1}$ to NB$_{60}$), while \textit{SBT\_unbalanced\_30s60n} is a dataset that contains fewer suspicious samples than normal ones (samples SB$_{1}$ to SB$_{30}$ and NB$_{1}$ to NB$_{60}$). To increase the number of samples, we applied a flipping procedure that consists of turn over horizontally each frame of the video sample, resulting in a clip where the actions happen in the opposite direction. For example, \textit{SBT\_balanced\_240} contains 240 samples (samples SB$_{1}$ to SB$_{60}$, NB$_{1}$ to NB$_{60}$, as well as the flipped versions of SB$_{1}$ to SB$_{60}$ and NB$_{1}$ to NB$_{60}$).
    
        \begin{table*}[ht!]
        \caption{Datasets description.}
        \label{tab:SamplesSelection}
        \resizebox{\textwidth}{!}{%
        \centering
        \begin{tabular}{ccc}
        \hline
        \textbf{Dataset} & \textbf{Suspicious samples} & \textbf{Normal samples} \\
        \hline
            SBT\_balanced\_60 & SB$_{1}$ to SB$_{30}$ & NB$_{1}$ to NB$_{30}$ \\
            SBT\_unabalanced\_30s60n & SB$_{1}$ to SB$_{30}$ & NB$_{1}$ to NB$_{60}$ \\
            SBT\_balanced\_120 & SB$_{1}$ to SB$_{60}$ & NB$_{1}$ to NB$_{60}$ \\
            SBT\_balanced\_120\_flip & Flipped versions of SB$_{1}$ to SB$_{60}$ & flipped versions of NB$_{1}$ to NB$_{60}$ \\
            SBT\_unbalanced\_60s120n & SB$_{1}$ to SB$_{60}$ & NB$_{1}$ to NB$_{60}$ + flipped versions of NB$_{1}$ to NB$_{60}$ \\
            SBT\_balanced\_240 & SB$_{1}$ to SB$_{60}$ + flipped versions of SB$_{1}$ to SB$_{60}$ & NB$_{1}$ to NB$_{60}$ + flipped versions of NB$_{1}$ to NB$_{60}$ \\
        \hline
        \end{tabular}
        }
        \end{table*}
    
    \subsection{Pre-Crime Behavior}
    \label{sec:PCB}
    
    To detect suspicious behavior, the proposed model must focus on what happens before a shoplifting crime is committed. For this purpose, we propose a new method to process surveillance videos. Before we explain our proposal, we introduce some concepts, which are listed below.
    
        \begin{itemize}
            \item \textbf{Strict Crime Moment (SCM).} In a surveillance video, and after being reviewed by a human, the SCM is the segment of video where a person commits shoplifting crime. This moment is the primary evidence to accuse a person of committing a crime.
            \item \textbf{Comprehensive Crime Moment (CCM).} it is the precise moment when an ordinary person can detect the suspect's intentions. He/she started to watch out to go unnoticed and looks for the best moment to commit the crime. Other CCM examples are unsuccessful attempts or reorder things to distract attention. If we isolate this moment, we can doubt the suspect in the video, but there is no clear evidence to know if the suspect steals something.
            \item \textbf{Crime Lapse (CL).} In a video, the CL is the entire segment where a crime takes place. If we remove the CL from the video, it will be impossible to determine that there is a criminal act in the video. The CCM supports the beginning of the CL. It is essential not to leave any trace of the crime to avoid biasing the training.
            \item \textbf{Pre-crime Behavior (PCB).} The PCB contains what happens from the first appearance of the suspect until the CCM begins. These samples have different sizes since each video shows many behaviors. We can find more than one CL per video. The next PCB will start where the previous CL ends and until the next CCM. The result is a video segment in which an ordinary person may not detect that a crime will occur, but we are sure that the sample comes from a video where criminal activity was present.
        \end{itemize}
        
        \begin{figure*}[ht!]
        \centering
            \includegraphics[width=12cm]{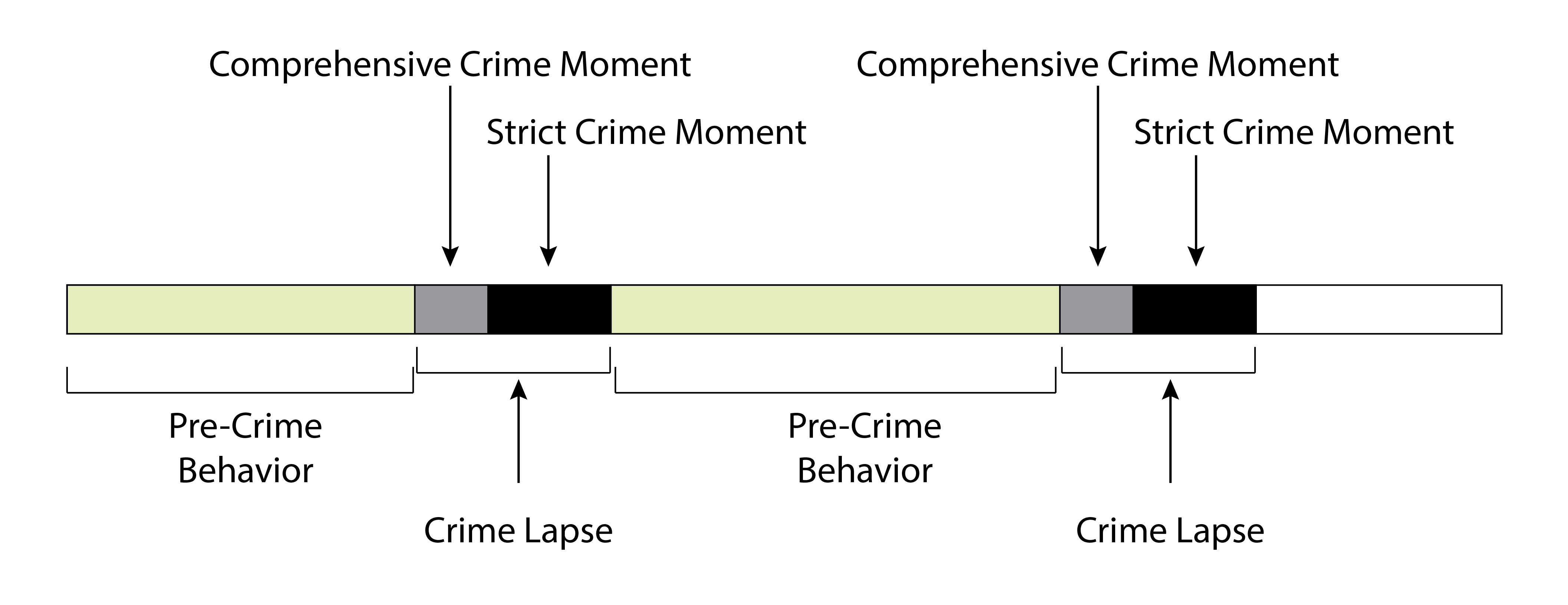}
            \caption{Graphical description of the concepts related to the proposed methodology for suspicious sample extraction.}
            \label{fig:Concepts}
        \end{figure*}
    
    Figure~\ref{fig:Concepts} graphically presents how these concepts interact in one video sample. The sample has two CL, and each one contains its corresponding SCM and CCM. From this video sample, we can extract two PCB training samples: from the beginning of the video to the first CCM, and from the end of the first SCM to the second CCM.
    
\begin{figure*}[ht!]
        \centering
            \includegraphics[width=\textwidth]{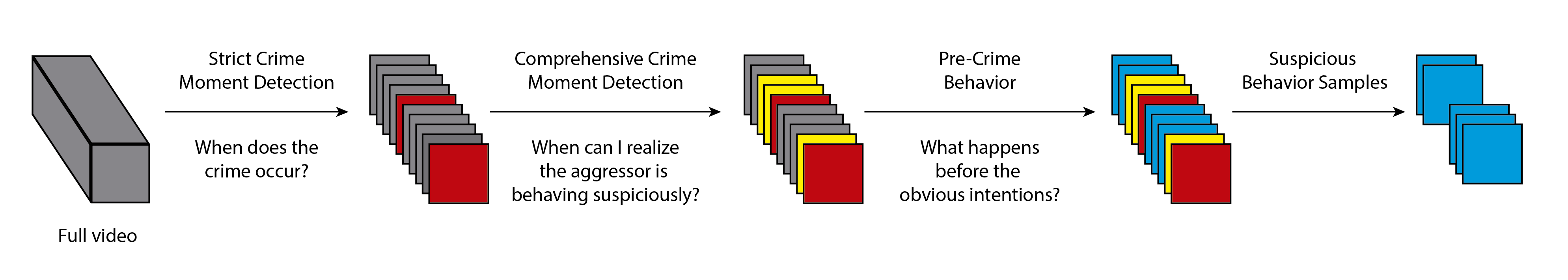}
            \caption{Graphical representation of the process for suspicious sample extraction.}
            \label{fig:samplesExtraction}
        \end{figure*}

    To extract the samples from the videos, we follow the process depicted in Fig.~\ref{fig:samplesExtraction}. Given a video that contains one or more shoplifting crimes, we identify the precise moment when the offense is committed. After that, we mark the different suspicious moments ---moments where a human observer doubts about what the person in the video is doing. Finally, we select the segment where the suspect people are preparing to commit the crime. These segments become the training samples for the Deep Learning (DL) model. 
    
\begin{figure*}[ht!]
        \centering
            \includegraphics[width=\textwidth]{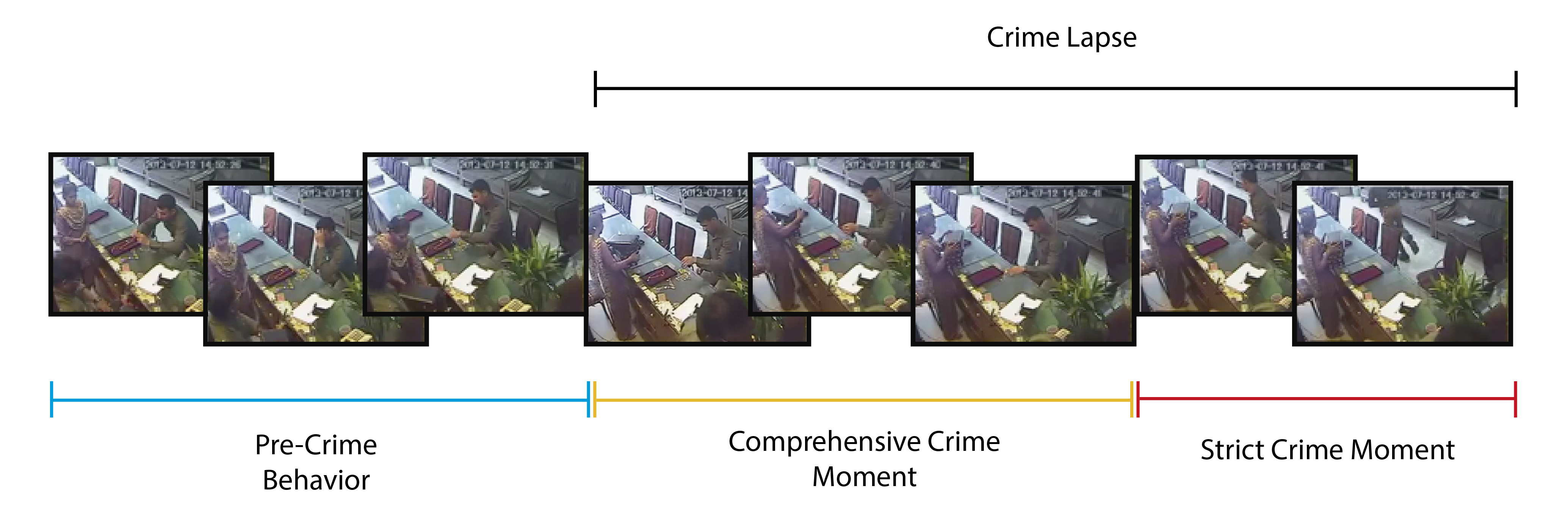}
            \caption{Video segmentation by critical moments.}
            \label{fig:Segmentation}
        \end{figure*}

    In a video sample, each moment has its information level importance (see Fig.~\ref{fig:Segmentation}). PCB has less information about the crime itself, but it allows us to analyze the suspect's normal-acting behavior in its first stage, even far from the target. CCM allows us to have a more precise idea about who may commit the crime, but it is not conclusive. Finally, SCM is the doubtless evidence about a person committing a shoplifting crime. If we remove SCM and CCM from the video, the result will be a video containing only people shopping, and there will be no suspicion or evidence if someone commits a crime. That is the importance of the accurate segmentation of the video. From where a Crime Lapse ends until the next SCM, there is new evidence about how a person behaves before committing a new shoplifting crime attempt.

    For experimentation purposes, we only use PCB segments. These segments lack specific criminal behavior and have no information about a transgression. We look to pattern an aggressor's behavior before trying to steal from a store.
    
    \subsection{3D Convolutional Neural Networks}
    \label{sec:3DCNN}

    We use a 3DCNN for feature extraction and classification. We choose a basic structure to explore the performance of the 3DCNN for suspicious behavior detection task. The architecture of the model consists of four Conv3D layers, two max-pooling layers, and two fully connected layers. As a default configuration, in the first pair of Conv3D layers, we apply 32 filters, and for the second pair, 64 filters. All kernels have a size of 3$\times$3$\times$3, and the model uses an Adam optimizer and cross-entropy for loss calculation. At the end of the model, it has two dense layers with 512 and 2 neurons, respectively. The output is binary, 1 for Suspicious Behavior and 0 for Normal Behavior. This architecture is selected because it has been used for similar applications~\cite{3DcnnImplementation}, and seems suitable as a first approach for behavior detection in surveillance videos.
    
    For handling the model training, we use Google Colaboratory~\cite{GoogleColab}. This free cloud tool allows to write and execute code in cells, runs from a browser, and uses a GPU to train deep learning models. We can upload the datasets to a storage service, link the files, prepare the training environment, and save considerable time during the model training, using a virtual GPU.

\section{Experiments and Results}
\label{sec:Experiments}

    3DCNN is a recent approach for Spatio-temporal analysis, showing a remarkable performance by processing videos in different areas, such as moving objects action recognition~\cite{HeObject3DCNN2017}, gesture recognition~\cite{ZhangGestures3DCNN2017} and action recognition~\cite{3dCNNHumanAction2013}. We decided to implement 3DCNN in a more challenging context, such as the search for patterns in criminal samples, which lack suspicious and illegal visual behavior. In this section, we present the proposed experiments and their results. 
    
    The initial experiment aims at exploring the impact of different values for the parameters of the system. The second experiment focuses on obtaining statistical support that the best configurations obtained from the first experiment are useful for further testing in different situations.

    \subsection{Exploration of configurations}
    \label{sub:Exploration}
    
        In this experiment, we explore different values for the parameters of the system. Given different values for the parameters, we estimate the changes in the response due to such configurations. The baseline training uses the most common values for this architecture, such as filters, kernel size, depth, and batch.  The rationale behind this first experiment is that by producing small variations on the input parameters, we expect to improve the model performance.
                    
        We consider an extensive set of parameters to generate the testing configurations and obtain a total of 22 configurations. For example, we consider a balanced dataset or flipped images. We use unbalanced datasets to simulate real environments where normal behavior is more likely to be present than suspicious ones. For these datasets, we use a sample ratio of 1:2; for each suspicious behavior sample, there are two normal behavior samples. The following is a short description of the nomenclature used to name the datasets so that the reader can understand the differences between each one of the datasets.
        
            \begin{itemize}
                \item \textbf{Balance.} The dataset has the same number of samples of each class. The values this parameter can take are balanced and unbalanced.
                \item \textbf{Ratio.} The proportion of samples of each class in the dataset. The different configurations for balanced sets include 60, 120 and 240 samples. For unbalanced datasets the ratio is 1:2 for suspicious~(s) and normal~(n) class, respectively, with a total of 90 and 180 samples.
                \item \textbf{Test size.} The percentage of the dataset intended to the testset. The possible percentage are 20, 30 or 40 percent.
                \item \textbf{Depth.} The number of consecutive frames used for a 3D convolution. The values this parameter can take are 10, 30 and 90.
                \item \textbf{Resolution.} The size of the input images. 160 $\times $120, 80 $\times $60, 40 $\times $30 or 32 $\times $24.
                \item \textbf{Flip.} If the word `flip' appear in the dataset name, the frames in the videos have turned over horizontally. 
            \end{itemize}
        
        By following the previous description, a dataset named \textit{SBT\_unbalanced\_60s120n\_ 30t\_30f\_40x30\_flip} refers to a dataset which has fewer samples of suspicious behavior than normal behavior, sixty and one hundred and twenty respectively. It destines thirty percent of the dataset for the test, and it uses thirty frames to perform a 3D convolution. Finally, the input images have a resolution of 40x30 pixels, and they were flipped horizontally. The tests focus on comparing different depths, test set sizes, the balance in the number of samples, which image resolution is optimal between time processing and image detail, and the data-augmentation technique of flip the images. The objective of the exploratory experiment is to find a suitable configuration to model suspicious behavior. As previously mentioned, we analyzed 22 configurations, which are tested in four different resolutions that run three times each.
        
        The depth sizes (number of consecutive frames) considered for the test are 10, 30 and 90. Table~\ref{tab:depth} shows the results of these runs with the four resolutions. Based on the results, using 10 and 30 frames achieves the best classification results, 69.4\% to 83.3\% and 69.4\% to 75\%, respectively. Table~\ref{tab:testSize} presents the results of the test set size comparison. We select values of 20\%, 30\%, and 40\% of the complete dataset for testing purposes. Although the first case (20\% test set size) uses more information to train, this proportion did not get the best results. It produced outcomes between 47.2\% and 72.2\%. The second case obtained results between 68.5\% and 75.9\%, and the third one between 61.1\% and 70.3\%. 30\% of the total dataset has the best results to define the test set size.
        
            \begin{table*}[ht!]
            \caption{Results of depth comparison.}
            \label{tab:depth}
            \centering
                \begin{tabular}{ccccc}
                \hline
                \textbf{} & \multicolumn{4}{c}{\textbf{Resolution}} \\
                \textbf{Dataset} & \textbf{32x24} & \textbf{40x30} & \textbf{80x60} & \textbf{160x120} \\ \hline
                SBT\_balanced\_60\_20t\_10f & \textbf{83.3\%} & 72.2\% & \textbf{77.7\%} & \textbf{69.4\%} \\
                SBT\_balanced\_60\_20t\_30f & 69.4\% & \textbf{75.0\%} & 69.4\% & \textbf{69.4\%} \\
                SBT\_balanced\_60\_20t\_90f & 69.4\% & 63.9\% & 61.1\% & 58.3\% \\
                \hline
                \end{tabular}
            \end{table*}
                    
            \begin{table}[ht!]
                \caption{Results of test set size comparison}
                \label{tab:testSize}
                \centering
                \resizebox{0.8\textwidth}{!}{
                \begin{tabular}{ccccc}
                \hline
                \multirow{2}{*}{\textbf{Dataset}} & \multicolumn{4}{c}{\textbf{Resolution}} \\ 
                 & \textbf{32x24} & \textbf{40x30} & \textbf{80x60} & \textbf{160x120} \\ \hline
                SBT\_balanced\_60\_20t\_10f & \textbf{72.2\%} & 72.2\% & 66.6\% & 47.2\% \\ 
                SBT\_balanced\_60\_30t\_10f & 68.5\% & \textbf{74.0\%} & \textbf{68.5\%} & \textbf{75.9\%} \\ 
                SBT\_balanced\_60\_40t\_10f & 63.9\% & 68.0\% & 61.1\% & 70.3\% \\ \hline
                \end{tabular}%
                }
            \end{table}
                    
        To deal with unbalanced training, we create three datasets with sixty normal samples, thirty suspicious ones and three different depths (Table~\ref{tab:unb}). We are aware that our model requires more samples to provide a better performance. However, in this test, the results reveal a similar performance, around 80\%, between 30 frames and 90 frames depth. 3DCNN can handle unbalanced datasets. The difference may relay in the training time of each depth.
        
            \begin{table}[ht!]
                \caption{Results of unbalanced dataset test}
                \label{tab:unb}
                \centering
                \resizebox{0.8\textwidth}{!}{%
                \begin{tabular}{ccccc}
                \hline
                \multirow{2}{*}{\textbf{Dataset}} & \multicolumn{4}{c}{\textbf{Resolution}} \\ 
                 & \textbf{32x24} & \textbf{40x30} & \textbf{80x60} & \textbf{160x120} \\ \hline
                SBT\_unbalanced\_30s60n\_30t\_10f & 66.6\% & 67.8\% & 68.8\% & 79.0\% \\ 
                SBT\_unbalanced\_30s60n\_30t\_30f & \textbf{69.1\%} & \textbf{65.4\%} & 80.2\% & \textbf{80.2\%} \\ 
                SBT\_unbalanced\_30s60n\_30t\_90f & \textbf{69.1\%} & 65.4\% & \textbf{81.4\%} & 66.6\% \\ \hline
                \end{tabular}%
                }
            \end{table}
            
        Data augmentation techniques are an option to take advantage of small datasets. For this reason, we test the model performance using original and flipped images in different runs. The used test set has a size of 30\% and 40\%. The tests throw accuracy results between 70\% and 80\% (Table~\ref{tab:flipped}). Therefore, we consider that both orientations can effectively be used as samples to train the model.
                
            \begin{table}[ht!]
                \caption{Results of flipped images comparison.}
                \label{tab:flipped}
                \centering 
                \resizebox{0.8\textwidth}{!}{%
                \begin{tabular}{ccccc}
                \hline
                \multirow{2}{*}{\textbf{Dataset}} & \multicolumn{4}{c}{\textbf{Resolution}} \\ 
                 & \textbf{32x24} & \textbf{40x30} & \textbf{80x60} & \textbf{160x120} \\ \hline
                SBT\_balanced\_120\_40t\_10f & 71.5\% & 71.5\% & 77.0\% & 77.0\% \\ 
                SBT\_balanced\_120\_40t\_10f\_flip & 73.6\% & 77.0\% & 83.3\% & 70.8\% \\ \hline
                SBT\_balanced\_120\_30t\_10f & 75.9\% & 71.3\% & 71.3\% & 79.6\% \\ 
                SBT\_balanced\_120\_30t\_10f\_flip & 76.8\% & 72.2\% & 81.5\% & 78.6\% \\ \hline
                \end{tabular}%
                }
            \end{table}
    
        Finally, we create datasets with balanced 240 samples and unbalanced 180 samples. Each type, balanced and unbalanced, combines three different depths and four resolutions, for a total of 24 datasets. Table~\ref{tab:resolution2} shows both the best result and the average result from three runs, for each resolution. It is essential to clarify better results by resolution may be achieved by using different depths. In this table, the value inside the parenthesis indicates the depth value.
        
        The presented results demonstrate that the best results are obtained through higher resolutions and using unbalanced datasets. Most of the results were achieved using depths of ten or thirty frames. The next experiments explore a more in-depth analysis of the best configurations, their performance, and statistical validation.
      
            \begin{table*}[tb]
                \centering
                \caption{\label{tab:resolution2}Best results comparison}
                \resizebox{\textwidth}{!}{%
                \begin{tabular}{ccccccccc}
                \hline
                \multirow{3}{*}{\textbf{Dataset}} & \multicolumn{8}{c}{\textbf{Resolution}} \\ 
                 & \multicolumn{2}{c}{\textbf{32x24}} & \multicolumn{2}{c}{\textbf{40x30}} & \multicolumn{2}{c}{\textbf{80x60}} & \multicolumn{2}{c}{\textbf{160x120}} \\ \cline{2-9} 
                 & Individual & Average & Individual & Average & Individual & Average & Individual & Average \\ \hline
                Balanced Dataset & 83.3\% (90f) & 75.9\% (30f) & 74.0\% (90f) & 73.4\% (90f) & 87.0\% (30f) & 79.6\% (10f) & 87.0\% (30f) & 79.6\% (10f) \\ 
                Unbalanced Dataset & 84.7\% (10f) & 77.7\% (30f) & 86.1\% (10f) & 76.3\% (30f) & 91.6\% (10f) & 87.0\% (10f) & 90.2\% (30f) & 86.1\% (30f) \\ \hline
                \end{tabular}%
                }
            \end{table*}
            
        \subsection{Statistical Validation}
        \label{sub:StatisticalValidation}
        
        Once the exploration tests end, we analyze the results to decide which parameters improve the classification and select configurations with the best performance. As a second experiment, the prominent configurations were run thirty times, using cross-validation, to give statistical support to the results that previously presented. For this experiment, the configurations train with the largest datasets we already create (SBT\_balanced\_240\_30t and SBT\_unbalanced\_60120\_30t). Complementing with cross-validation, we use ten folds of the dataset for train and test.
                
        From previous results, four configurations were trained with 240 balanced-samples and 180 unbalanced-samples datasets. Fixed parameters were 100 epochs for training, 70\% samples for training and 30\% for testing, both datasets use the original and the flipped images. The ratio and number of samples per class can be inferred from the balance parameter (see section~\ref{sub:Exploration}, balance and ratio). For this test, we perform thirty runs per configuration and use ten dataset folds for cross-validation.   
        
        Table~\ref{tab:30runs} presents average accuracy and the standard deviation of each configuration tested. Most of the results are around 70\% accuracy. There is not significative deviation on each training group. The results seem very similar between them. We analyze the confusion matrices to search for biased results. Although we find cases were the classification results are biased to a particular class, we discover good results. 
        
            \begin{table}[ht!]
                \caption{Thirty runs training results.}
                \label{tab:30runs}
                \centering
                \resizebox{0.8\textwidth}{!}{%
                \begin{tabular}{cccc}
                \hline
                \textbf{Resolution} & \textbf{Dataset} & \textbf{Avg Accuracy} & \textbf{Std Deviation} \\ \hline
                \multirow{4}{*}{160x120} & unb\_60s120n\_30t\_10f & 75.7\% & 0.0638 \\ 
                 & unb\_60s120n\_30t\_30f & 73.9\% & 0.0543 \\ 
                 & bal\_240\_30t\_10f & 73.1\% & 0.0661 \\ 
                 & bal\_240\_30t\_30f & 71.6\% & 0.0999 \\ \hline
                \multirow{4}{*}{80x60} & unb\_60s120n\_30t\_10f & 75.0\% & 0.0689 \\ 
                 & unb\_60s120n\_30t\_30f & 74.8\% & 0.0500 \\ 
                 & bal\_240\_30t\_10f & 73.0\% & 0.0717 \\ 
                 & bal\_240\_30t\_30f & 73.6\% & 0.0821 \\ \hline
                \multirow{4}{*}{40x30} & unb\_60s120n\_30t\_10f & 68.7\% & 0.0569 \\ 
                 & unb\_60s120n\_30t\_30f & 69.1\% & 0.0576 \\ 
                 & bal\_240\_30t\_10f & 71.8\% & 0.0468 \\ 
                 & bal\_240\_30t\_30f & 71.9\% & 0.0555 \\ \hline
                \multirow{4}{*}{32x24} & unb\_60s120n\_30t\_10f & 69.4\% & 0.0686 \\ 
                 & unb\_60s120n\_30t\_30f & 71.6\% & 0.0533 \\ 
                 & bal\_240\_30t\_10f & 70.3\% & 0.0476 \\ 
                 & bal\_240\_30t\_30f & 70.1\% & 0.0574 \\ \hline
                \end{tabular}%
                }
            \end{table}
        
        In this investigation, the 80x60 resolution has the best results in suspicious behavior detection task. It achieves accuracy rates above 85\% both balanced and unbalanced datasets, preferably with ten frames depth. The best result in a single run is 92.50\% of accuracy. This performance was obtained in the thirtieth run, using the unbalanced dataset and ten frames depth. After 30 runs, the model obtain an average accuracy of 75\%.
                
        Table~\ref{tab:confusionMatrix} exhibits the best results and their confusion matrices. Even in the confusion matrices, the accuracy per class is above 90\% for suspicious-behavior class and around 80\% for normal-behavior class.
        
            \begin{table}[htb]
                \centering
                \caption{Confusion Matrix of best results}
                \label{tab:confusionMatrix}
                \resizebox{0.5\textwidth}{!}{%
                \begin{tabular}{cccc}
                \multicolumn{4}{l}{Dataset: \textbf{unb\_60s120n\_30t\_10f\_80x60}} \\ 
                \multicolumn{4}{l}{Accuracy: \textbf{92.5\%}} \\ \hline
                \multicolumn{1}{c|}{} & \multicolumn{1}{c}{\textbf{Suspicious}} & \multicolumn{1}{c}{\textbf{Normal}} & \multicolumn{1}{c}{\textbf{Accuracy}} \\ \hline
                \multicolumn{1}{c|}{\textbf{Suspicious}} & \multicolumn{1}{c}{18} & \multicolumn{1}{c}{0} & \multicolumn{1}{c}{100\%} \\ 
                \multicolumn{1}{c|}{\textbf{Normal}} & \multicolumn{1}{c}{4} & \multicolumn{1}{c}{32} & \multicolumn{1}{c}{88.9\%} \\ \hline
                \multicolumn{4}{l}{} \\ 
                \multicolumn{4}{l}{Dataset: \textbf{bal\_240\_30t\_10f\_80x60}} \\ 
                \multicolumn{4}{l}{Accuracy: \textbf{91.6\%}} \\ \hline
                \multicolumn{1}{c|}{\textbf{}} & \multicolumn{1}{c}{\textbf{Suspicious}} & \multicolumn{1}{c}{\textbf{Normal}} & \multicolumn{1}{c}{\textbf{Accuracy}} \\ \hline
                \multicolumn{1}{c|}{\textbf{Suspicious}} & \multicolumn{1}{c}{36} & \multicolumn{1}{c}{0} & \multicolumn{1}{c}{100\%} \\ 
                \multicolumn{1}{c|}{\textbf{Normal}} & \multicolumn{1}{c}{6} & \multicolumn{1}{c}{30} & \multicolumn{1}{c}{83.3\%} \\ \hline
                \multicolumn{4}{l}{} \\ 
                \multicolumn{4}{l}{Dataset: \textbf{unb\_60s120n\_30t\_10f\_80x60}} \\ 
                \multicolumn{4}{l}{Accuracy: \textbf{90.7\%}} \\ \hline
                \multicolumn{1}{c|}{\textbf{}} & \multicolumn{1}{c}{\textbf{Suspicious}} & \multicolumn{1}{c}{\textbf{Normal}} & \multicolumn{1}{c}{\textbf{Accuracy}} \\ \hline
                \multicolumn{1}{c|}{\textbf{Suspicious}} & \multicolumn{1}{c}{18} & \multicolumn{1}{c}{0} & \multicolumn{1}{c}{100\%} \\ 
                \multicolumn{1}{c|}{\textbf{Normal}} & \multicolumn{1}{c}{5} & \multicolumn{1}{c}{31} & \multicolumn{1}{c}{86.0\%} \\ \hline
                \multicolumn{4}{l}{} \\ 
                \multicolumn{4}{l}{Dataset: \textbf{bal\_240\_30t\_10f\_80x60}} \\ 
                \multicolumn{4}{l}{Accuracy: \textbf{90.2\%}} \\ \hline
                \multicolumn{1}{c|}{\textbf{}} & \multicolumn{1}{c}{\textbf{Suspicious}} & \multicolumn{1}{c}{\textbf{Normal}} & \multicolumn{1}{c}{\textbf{Accuracy}} \\ \hline
                \multicolumn{1}{c|}{\textbf{Suspicious}} & \multicolumn{1}{c}{36} & \multicolumn{1}{c}{0} & \multicolumn{1}{c}{100\%} \\ 
                \multicolumn{1}{c|}{\textbf{Normal}} & \multicolumn{1}{c}{7} & \multicolumn{1}{c}{29} & \multicolumn{1}{c}{80.6\%} \\ \hline
                \end{tabular}
                }
            \end{table}
            
            \begin{table}[htb]
                \caption{Comparison between base model and the proposed one.}
                \label{tab:FinalComparisonWithMatrx}
                \centering
                \resizebox{0.8\textwidth}{!}{%
                \begin{tabular}{ccccccccc}
                \hline
                \multicolumn{1}{c}{\multirow{2}{*}{}} & \multicolumn{4}{c}{\textbf{Balanced}} & \multicolumn{4}{c}{\textbf{Unbalanced}} \\ 
                \multicolumn{1}{c}{} & \multicolumn{2}{c}{\textbf{Base model}} & \multicolumn{2}{c}{\textbf{Proposed}} & \multicolumn{2}{c}{\textbf{Base model}} & \multicolumn{2}{c}{\textbf{Proposed}} \\ \hline
                \multicolumn{1}{c}{\textbf{Avg Acc}} & \multicolumn{2}{c}{71.7\%} & \multicolumn{2}{c}{\textbf{73.0\%}} & \multicolumn{2}{c}{70.5\%} & \multicolumn{2}{c}{\textbf{75.0\%}} \\ 
                \multicolumn{1}{c}{\textbf{Std Dev}} & \multicolumn{2}{c}{0.06} & \multicolumn{2}{c}{0.07} & \multicolumn{2}{c}{0.05} & \multicolumn{2}{c}{0.07} \\ 
                \multicolumn{1}{c}{\textbf{Best Result}} & \multicolumn{2}{c}{81.9\%} & \multicolumn{2}{c}{91.6\%} & \multicolumn{2}{c}{88.8\%} & \multicolumn{2}{c}{92.5\%} \\ \hline
                \multicolumn{1}{l}{} & \multicolumn{1}{l}{} & \multicolumn{1}{l}{} & \multicolumn{1}{l}{} & \multicolumn{1}{l}{} & \multicolumn{1}{l}{} & \multicolumn{1}{l}{} & \multicolumn{1}{l}{} & \multicolumn{1}{l}{} \\ 
                \multicolumn{1}{l}{} & \multicolumn{1}{l}{} & \multicolumn{3}{c}{\textbf{Base Model}} & \multicolumn{1}{l}{} & \multicolumn{3}{c}{\textbf{Proposed}}  \\ 
                \multirow{3}{*}{\textbf{Balanced}} &  & \multicolumn{1}{c}{} & \multicolumn{1}{|c}{Susp} & \multicolumn{1}{c}{Norm} &  & \multicolumn{1}{c}{} & \multicolumn{1}{|c}{Susp} & \multicolumn{1}{c}{Norm} \\ \cline{3-5} \cline{7-9} 
                 & \multicolumn{1}{c}{} & \multicolumn{1}{c|}{Susp} & \multicolumn{1}{c}{34} & \multicolumn{1}{c}{2} & \multicolumn{1}{c}{} & \multicolumn{1}{c|}{Susp} & \multicolumn{1}{c}{36} & \multicolumn{1}{c}{0} \\ 
                 & \multicolumn{1}{c}{} & \multicolumn{1}{c|}{Norm} & \multicolumn{1}{c}{11} & \multicolumn{1}{c}{25} & \multicolumn{1}{c}{} & \multicolumn{1}{c|}{Norm} & \multicolumn{1}{c}{6} & \multicolumn{1}{c}{30} \\ \cline{3-5} \cline{7-9} 
                 &  &  &  &  &  &  &  &  \\ 
                \multirow{3}{*}{\textbf{Unbalanced}} &  & \multicolumn{1}{c}{} & \multicolumn{1}{|c}{Susp} & \multicolumn{1}{c}{Norm} &  & \multicolumn{1}{c}{} & \multicolumn{1}{|c}{Susp} & \multicolumn{1}{c}{Norm} \\ \cline{3-5} \cline{7-9} 
                 & \multicolumn{1}{c}{} & \multicolumn{1}{c|}{Susp} & \multicolumn{1}{c}{16} & \multicolumn{1}{c}{2} & \multicolumn{1}{c}{} & \multicolumn{1}{c|}{Susp} & \multicolumn{1}{c}{18} & \multicolumn{1}{c}{0} \\ 
                 & \multicolumn{1}{c}{} & \multicolumn{1}{c|}{Norm} & \multicolumn{1}{c}{4} & \multicolumn{1}{c}{32} & \multicolumn{1}{c}{} & \multicolumn{1}{c|}{Norm} & \multicolumn{1}{c}{4} & \multicolumn{1}{c}{32} \\ \cline{3-5} \cline{7-9} 
                \end{tabular}
                }
            \end{table}
        
    \subsection{Discussion}
    \label{sub:Discussion}
    
        As the first experiment in this work, we select a 3D Convolutional Neural Network with a basic configuration as a base model, and then we perform a parameter tunning, searching for network model improvement.
    
        From the parameter exploration, we found that 80x60 and 160x120 resolutions deliver better results than a commonly used low resolution or. This experiment was limited to a maximum resolution of 160x120 due to processing resources.
    
        Another significant aspect is the "depth" parameter. This parameter describes the number of consecutive frames used to perform the 3D convolution. After testing different values, we observed that low values, between 10 and 30 frames, have a good relationship between image detail and processing time. These two factors impact the network model training and the correct classification of the samples.
    
        Also, the proposed model can correctly handle flipped images and unbalanced datasets. We performed a more realistic simulation where the dataset has more normal-behavior samples than suspicious-behavior examples. The unbalanced datasets were also correctly classified.
    
        For the second experiment, we use the configurations with the best performance and test them with bigger datasets. We performed 30 runs for each configuration, applying cross-validation, with 10 and 30 frames values for depth and using a 240-samples balanced dataset and a 180-samples unbalanced dataset.
    
        From this experimentation, we found that 80x60 resolution reports better accuracy for the four scenarios we test. Table~\ref{tab:30runs} presents the average accuracy for each configuration. The average performance for the four scenarios in 80x60 resolution is 74.1\%, while the 160x120 resolution obtains 73.5\%. Also, in a single training, 80x60 resolution performance achieves over 90\%, 92.5\% for a balanced dataset and 91.6\% for the unbalanced dataset.
    
        Finally, when comparing the base model against the proposed one (Table~\ref{tab:FinalComparisonWithMatrx}), we obtained that our model is capable of improving the classification results by 1.3\% and 4.5\% on average for balanced and unbalanced datasets, respectively. In the best-single-training comparison, the proposed architecture exceeds 90\% accuracy in both cases. The confusion matrices show that for both balanced and unbalanced datasets, the proposed architecture successfully classifies 100\% of suspicious samples and obtains a low number of false positives from normal-behavior samples.
    
    \subsection{Processing Time}
    \label{ProcessingTime}
        
        As mentioned before, we use Google Colaboratory to perform the experiments. This tool is based on Jupyter Notebooks and allows the free use of a GPU. The speed of each training depends on the tool demand. Most of the network trainings end in less than an hour, but a higher GPU demand may impact the training time. We are not able to establish a relationship between resolutions and training time, but we have an approximate correlation between different depths. 
        
        Table~\ref{tab:TrainingTime} shows the average training time of the final tests, running one hundred epochs, described in section~\ref{sub:Exploration} (accuracy results of these experiments are shown in Table~\ref{tab:resolution2}). Comparing the training time from using 10-frames against 30-frames, it increases approximately three times in the second training. We find the same increase's relation when comparing trainings with 30-frames and 90-frames. Some ninety-frames trainings, with a hundred epochs and high resolution, have reached a duration of up to four hours. Training duration is an essential factor due to the size of the used dataset is considerably small.
    
        Another point to consider is the system's accuracy against the training time. Although the training time increases approximately three times, using the same number of epochs, the accuracy is usually lower when using 90-frames depth, in most of the cases. In some instances, we get a higher precision using 90-frames, but accuracy reached by smaller depths was not far from the best one, and the training time was considerably lower.
        
            \begin{table}[ht!]
                \centering
                \caption{Average training times in seconds comparison between different depths and resolutions.}
                \label{tab:TrainingTime}
                \resizebox{0.8\textwidth}{!}{%
                \begin{tabular}{c|c|rrrr}
                \hline
                \multicolumn{2}{c}{\multirow{2}{*}{\textbf{Dataset}}} & \multicolumn{4}{c}{\textbf{Resolution}} \\ 
                \multicolumn{2}{c}{} & \multicolumn{1}{c}{\textbf{32x24}} & \multicolumn{1}{c}{\textbf{40x30}} & \multicolumn{1}{c}{\textbf{80x60}} & \multicolumn{1}{c}{\textbf{160x120}} \\ \hline
                \multirow{3}{*}{\textbf{SBT\_unbalanced\_60120...}} & \textbf{10f} & 96 & 126 & 369 & 1,356 \\ 
                 & \textbf{30f} & 196 & 279 & 1,027 & 3,918 \\ 
                 & \textbf{90f}& 518 & 758 & 2,929 & 11,655 \\ \hline
                \multirow{3}{*}{\textbf{SBT\_balanced\_240...}} & \textbf{10f} & 118 & 157 & 475 & 1,714 \\ 
                 & \textbf{30f} & 257 & 364 & 1,304 & 4,952 \\ 
                 & \textbf{90f} & 688 & 1,011 & 3,879 & 15,415 \\ \hline
                \end{tabular}%
                }
            \end{table}

\section{Conclusion}
\label{sec:Conclusion}

    For this work, we focus on the behavior performed by a person before committing a shoplifting crime. The neural network model identifies the previous conduct, looking for suspicious behavior and not to recognize the crime itself. This behavior analysis is the principal reason why we remove the committed crime segment from the video samples, to allow the artificial model to focus on decisive conduct and not in the offense. We implement a 3D Convolutional Neural Network due to its capability to obtain abstract features from signals and images, based on previous approaches for action recognition and movement detection. 
    
    Based on the results obtained from the conducted experimentation, a 75\% accuracy in suspicious behavior detection, we can state that it is possible to model the suspicious behavior of a person in the shoplifting context. Through the presented experimentation, we found which parameters fit better for behavior analysis, particularly for the shoplifting context. We explore different parameters and configurations, and in the end, we compare our results against a reference 3D Convolutional architecture. The proposed model demonstrates a better performance with balanced and unbalanced datastes and using the particular configuration obtained from previous experiments.
    
    The final intention of this experimentation is the development of a tool capable of supporting the surveillance staff, presenting visual behavioral cues, and this work is a first step to achieve the mentioned goal. From this point, we will explore different aspects that will contribute to the project development, such as bigger datasets, adding more criminal contexts that present suspicious behavior and real-time tests.

    \subsection{Future Work}
    \label{sub:FutureWork}
    
        For these experiments, we use a selected number of videos from the UCF-Crimes dataset. To test in a more realistic simulation, we have to increase the number of samples, preferably the normal-behavior ones, to create a bigger sample-unbalance between classes.
    
        Another interesting aspect of the developing of this project is to expand our behavior detection model to other contexts. It exists many situations where we can find suspicious behavior, such as stealing, arson intents, burglary, among others. We will gather videos of different contexts to strengthen the capability to detect suspicious behavior.
    
\bibliographystyle{unsrt}
\bibliography{references}

\end{document}